# Fast 3D Surrogate Modeling for Data Center Thermal Management

Soumyendu Sarkar[*,†], Antonio Guillen-Perez[†], Zachariah J Carmichael[†], Avisek Naug[†], Refik Mert Cam[†], Vineet Gundecha, Ashwin Ramesh Babu, Sahand Ghorbanpour, Ricardo Luna Gutierrez

Hewlett Packard Enterprise
820 N McCarthy Blvd,
Milpitas, CA 95035, USA
{soumyendu.sarkar, antonio.guillen, zach.carmichael, avisek.naug, refik-mert.cam,
vineet.gundecha, ashwin.ramesh-babu, sahand.ghorbanpour, rluna}@hpe.com

## Abstract

Reducing energy consumption and carbon emissions in data centers by enabling real-time temperature prediction is critical for sustainability and operational efficiency. Achieving this requires accurate modeling of the 3D temperature field to capture airflow dynamics and thermal interactions under varying operating conditions. Traditional thermal CFD solvers, while accurate, are computationally expensive and require expert-crafted meshes and boundary conditions, making them impractical for real-time use. To address these limitations, we develop a vision-based surrogate modeling framework that operates directly on a 3D voxelized representation of the data center, incorporating server workloads, fan speeds, and HVAC temperature set points. We evaluate multiple architectures, including 3D CNN U-Net variants, a 3D Fourier Neural Operator, and 3D vision transformers, to map these thermal inputs to high-fidelity heat maps. Our results show that the surrogate models generalize across data center configurations and significantly speed up computations (20,000x), from hundreds of milliseconds to hours. This fast and accurate estimation of hot spots and temperature distribution enables real-time cooling control and workload redistribution, leading to substantial energy savings ( 7%) and reduced carbon footprint.

## Introduction

Machine learning (ML) surrogate models have become efficient alternatives to conventional computational techniques. These surrogate models learn from data and deliver quick and effective results by simplifying complex, resource-intensive simulations. A key application area has been substituting the computationally intensive heat flow models typically derived from Computational Fluid Dynamics (CFD) simulations in data center operations. Traditional CFD solvers, though accurate, require expert-crafted meshes, boundary condition tuning, and domain expertise, making them impractical for real-time decision support.

The complexity and scale of data centers, which house a range of computing devices, including servers, network equipment, and advanced cooling systems, make them challenging to model. Traditioal CFD simulations in this domain often take several days or weeks to complete. This long duration conflicts with the real-time decision-making needs of data centers, where quick actions improve energy efficiency, performance, and environmental impact.

This paper introduces a novel thermal data modeling approach for building accurate CFD surrogates for data centers (CFDDC). The proposed approach operates directly on a voxelized representation that captures a data center's complex three-dimensional geometry of heat sources, cooling system components, power consumption, and airflow. It helps pinpoint hot spots and analyze temperature distributions, achieving several orders of magnitude speed-up compared to CFD approaches while maintaining high accuracy. The goal is to equip data centers with a tool for faster optimization of workload distribution, cooling adjustment, and server placement, with an emphasis on energy efficiency and sustainability.

While we leverage established architectures such as 3D U-Net, Swin UNETR and Fourier Neural Operator (FNO), our novel contribution lies in the unique design of the input data channels, that are adapted to encode thermal dynamics, allowing the model to efficiently predict temperature distributions.

The main contributions of this research are:

- We propose a fast CFD surrogate for the thermal modeling of data centers to predict hot spots. The approach scales well with the size of the data centers and enables downstream optimization tasks that contribute to sustainability.

- We introduce a novel 3D thermal modeling technique, where the input channels represent the thermal attributes like heat sources, flows, vents, and controls while the output represents temperature in 3D space. This modeling works both for 3D CNN architectures like U-Net variants, 3D FNOs and 3D vision transformers, establishing its robustness. To the best of our knowledge, this is the first approach for this application.

---

[*]Corresponding author.
[†]These authors contributed equally.

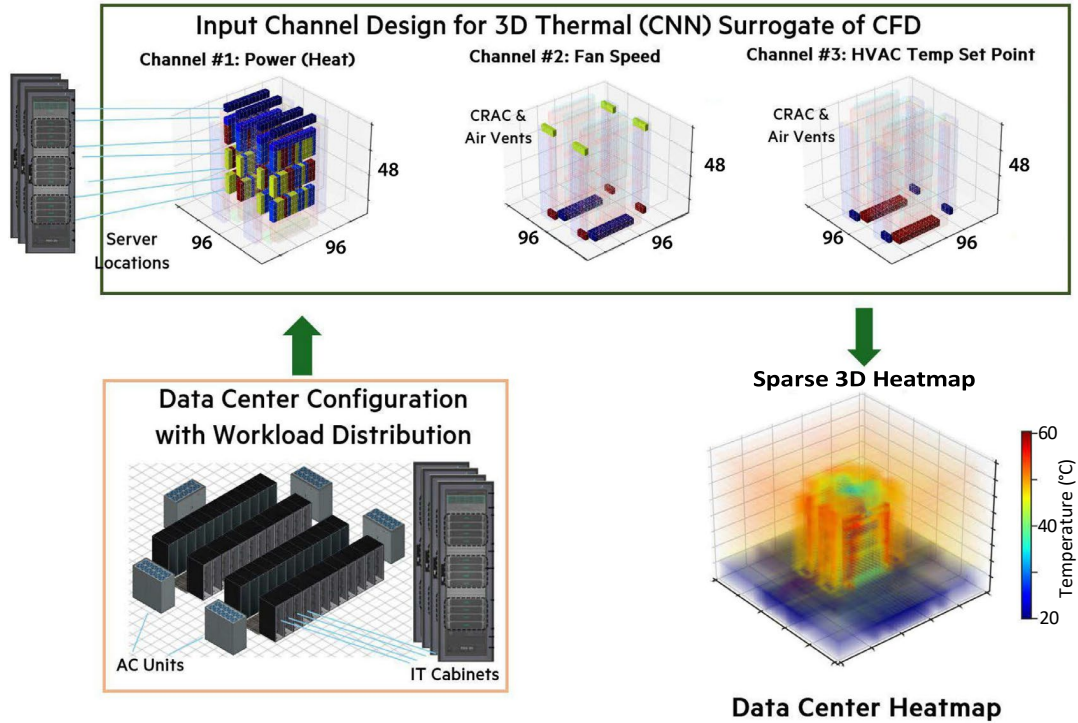

Figure 1: Novel input channels of the 3D Vision model for CFD surrogate.

The following sections detail our approach, its implementation, and its impact on sustainable data center operations.

## Background and Related Work

As data centers aim for energy efficiency, ML surrogates are increasingly used to replace CFD simulations. These data-driven models are preferred over CFD simulations for the speed of execution, enabling optimization of data centers with varying configurations and adjustments to workloads, racks, IT equipment, Computer Room Air Conditioning (CRAC) unit locations, cold aisle containment, etc.

For DC server thermal performance analysis, (Ilager, Ramamohanarao, and Buyya 2020) experimented with XGBoost, Linear Regression, and a fully connected network, finding XGBoost yielded the best results. (Lin et al. 2022) employed six machine learning algorithms: SVR, GPR, XGBoost, LightGBM, ANN, and LSTM for similar problems. XGBoost and LightGBM outperformed the others. In (Morozova et al. 2022; Jin et al. 2023; Athavale, Yoda, and Joshi 2019) surrogate models were developed using ANN, SVR, GBR, and GPR and reduced order approaches to predict steady state fluid variables with varying configurations and boundary conditions in a ventilated room. Most of these approaches are restricted to 1D time series modeling, while we are interested in 3D steady state thermal models. In (Asgari et al. 2021) a hybrid model was designed to predict the transient thermal behavior of server CPUs and the cold chamber in a modular DC with a row-based cooling architecture. These approaches have focused exclusively on estimating temperature at a single point in space over time. In this regard, over the last few years, several novel ML techniques have been created that focus on solving differential equations by mapping them directly to the solution space using neural networks (Li et al. 2020; Xiao et al. 2023; Guo, Li, and Iorio 2016). They are orders of magnitude faster than solving Partial Differential Equations (PDEs) using CFD approaches. This work focuses on 3D steady-state temperature heatmaps rather than flow variables.

Recent advances in surrogate modeling for computational fluid dynamics (CFD) have highlighted the limitations of conventional deep learning architectures in efficiently capturing global dependencies within volumetric data for thermal prediction tasks. Traditional convolutional models, while effective at learning local spatial relationships, often struggle with generalization to new configurations and require extensive retraining to handle changes in boundary conditions or physical parameters (Li et al. 2020; Xiao et al. 2023).

The **Fourier Neural Operator** (FNO) addresses these challenges by parameterizing solution spaces in the Fourier domain, thereby learning mappings between infinite-dimensional function spaces associated with PDE solutions. This approach enables global receptive fields and rapid inference, making it particularly well-suited for real-time prediction scenarios in large-scale data center thermal modeling (Li et al. 2020; Guibas et al. 2021). The use of spectral learning allows the FNO to capture long-range spatial correlations and complex physical interactions, providing superior performance, resolution invariance, and faster convergence compared to traditional neural network architec-tures (Li et al. 2020; Peng et al. 2024).

The primary motivation for employing FNO in our CFD surrogate framework is its proven ability to achieve state-of-the-art accuracy and generalization, while offering up to three orders of magnitude speedup for typical physical simulations (Li et al. 2020, 2025). This empowers data centers with tools for rapid optimization and adaptive control, directly supporting sustainability and energy efficiency objectives.

In our work, we leveraged state-of-the-art 3D CNNs and a 3D FNO for surrogate model design, since CFD tools produce volumetric 3D images of thermal flow. We gathered training data through appropriate sampling across all input space dimensions using $6SigmaDCX$ (6SigmaDCX — Future Facilities 2023). We selected various 3D CNN models for training, such as U-Net (Ronneberger, Fischer, and Brox 2015), Residual-UNet (Lee et al. 2017), UNet++ (Zhou et al. 2019), Residual-UNetSE (Toubal, Duan, and Yang 2020), Swin UNETR (Hatamizadeh et al. 2022) and we also included a 3D FNO to efficiently capture non-local spatial correlations in the temperature field (Li et al. 2020). Our data processing approach coupled with these models indicate that the thermal surrogates achieve high prediction accuracy, making them suitable alternatives to CFD tools.

## Methodology

In our novel design for input channels, the three channels encode the three distinct thermodynamic attributes. The first channel encodes the normalized volumes for heat sources (heat generators). The second channel encodes the flow rates at the 2D (single voxel height) slices for the intake and outtake vents and the vent from the AC unit to the pressurized underfloor of data centers (heat sinks). The third channel encodes the AC unit control as temperature set point(s) with appropriate location distribution in the 3D data center space (heat sinks). The modeling of these three channels is vital to enable the 3D CNN to estimate the temperature heat map.

### Data Center Models

We used $6SigmaDCX$ (6SigmaDCX — Future Facilities 2023) to design detailed models of data centers in various configurations (Figure 2). Each cabinet consists of three segments (splits) with four HPE DL380 servers (see Figure 1).

### CFD Data Acquisition

To generate the dataset required for our study, we employed the structured sampling methodology known as Latin Hypercube Sampling (LHS) (McKay, Beckman, and Conover 1979). The group of servers (splits) can have three possible workload utilization levels: $25\%$, representing low demand; $60\%$, representing medium demand; and $90\%$, representing high demand. By employing this sampling strategy, we ensured the generation of a robust and diverse dataset, capturing not only typical operational scenarios but also conditions indicative of stress or imbalance, leading to temperature differentials and hotspots at a high sample efficiency. These percentages can be changed with alternate workload distribution strategies.

During the data collection phase, close to 10000 CFD samples were generated across various data center configurations, with simulation times ranging from 11 minutes for simpler setups to 35 minutes for the most complex/biggest ones. Figure 3 outlines these configurations, including simulation times and the split (train or test). Seven unique configurations were created for training, spanning a spectrum of sizes to help the model learn to generalize heat flow patterns to larger and more diverse data center environments.

We want to highlight here that collecting data on dimensions higher than 2 for complex PDE experiments, especially for a large data center, related to ML, has turned out to be a difficult task in other domains too (See section 6 in (Li et al. 2020).) and is still a work in progress in different research projects.

The data acquisition process was automated using $6SigmaCommander^{TM}$. Based on the above-sampled workload and setpoints, we ran the CFDs. After the CFD thermal simulation, we stored the cabinet/server/ACU specifications (input data, $X$) and the resulting 3D temperature mesh (output data, $Y$) from the CFD solver.

### 3D Input Data Processing

The major innovation in our approach is not the structural design of the neural networks but how we designed the input data for thermal modeling. We preprocess our CFD-generated data into structured 3D tensors that encapsulate critical thermal attributes of data centers, thus enabling the network to perform spatially and thermodynamically relevant feature extraction more effectively.

The surrogate model takes a 3D voxel tensor as input, which encodes the geometry and thermal metrics of a data center. These metrics include power of IT components, Computer Room Air Conditioning (CRAC) units set point, and CRAC units fan speed, each occupying one channel of the input. The power serves as a proxy for heat generators, while the CRAC set point and fan speed are proxies for cooling and airflow. The IT components represented include cabinets, cabinet equipment, CRAC and its components (supplies and returns), and slotted grilles. In the voxel tensor, each component is placed according to its exact 3D position and geometry as defined in $6SigmaDCX$: a coordinate $\langle x, y, z \rangle$, an orientation $\theta \in [0, 360)$, and dimensions $\langle w, h, d \rangle$. Each component is a cuboid with width $w$, height $h$, and depth $d$. The $6SigmaDCX$ axes are measured in meters and must be quantized for a voxel grid. We select to use approximately 6.6 voxels per meter, resulting in a quantized room size of $128 \times 32 \times 128$ voxels, in contrast to the original room size of $25 \times 4 \times 25$ meters. These give us a resolution of 15 cm per voxel. We elect to use these dimensions as (1) they do not over-quantize (destroying information, e.g., reducing the volume of a component to zero), (2) the memory requirements and inference times are not excessive, and (3) they are more compatible with the down- and up-sampling dimensions of the 3D CNNs. For all channels, the associated metric is volume-normalized, that is, a component with a metric value of $v$ is assigned to each of the voxels for that component as $v/(whd)$. Intuitively, this encodes the concentration/dispersion of that metric; for instance, a large component will have heat spread over a larger area but at a lower level than a small component with equal power utilization. In summary, each channel contains the following information and geometrically depicted in Figure 1. For more detailed information on the rationale behind choosing the $128 \times 32 \times 128$ resolution, refer to the supplementary document.

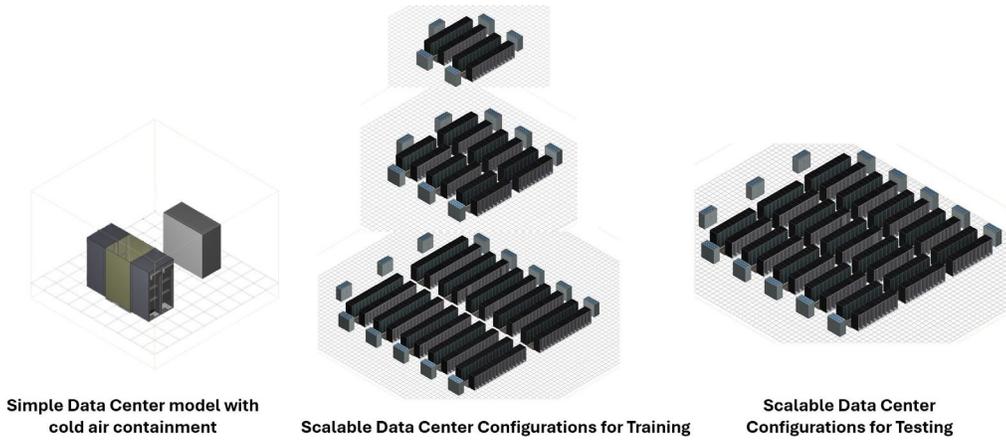

Figure 2: Data center model configuration of various scales for which the CFD heat map was generated using $6SigmaDCX$ tool to train and test (for different configurations).

| Data Center Configuration | | | | |
|---|---|---|---|---|
| Train/Test | Train | Train | Train | Train |
| CFD Simulation Execution Time | 11 mins | 11 mins | 12 mins | 21 mins |
| Data Center Configuration | | | | |
| Train/Test | Train | Train | Train | Test |
| CFD Simulation Execution Time | 23 mins | 22 mins | 35 mins | 35 mins |
| ML Surrogate Execution Time | | | ~0.1 s | ~0.1 s |

Figure 3: Data centers configurations designed

- **Power**: This channel encodes the volume-normalized power utilization in kW of the servers, switches, and storage arrays within each cabinet. The power is set to 0 kW in all other voxels without components.
- **CRAC Set Point**: This channel encodes the volume-normalized and negated CRAC set point in Celsius (C) of the CRAC supplies and slotted grilles, such that higher values are cooler. The slotted grilles are included to indicate the target temperature of cool air that is flowing from the floor. The value is set to 0 in all other voxels without components.
- **CRAC Fan Speed**: This channel encodes the volume-normalized CRAC fan speed as a percentage of the CRAC supplies, CRAC returns, and slotted grilles. The fan speed is set to 0% in all other voxels without components.

These channels are concatenated to create each input tensor $x \in \mathrm{R}^{C \times D \times H \times W}$ where $C$ is the number of channels (three in our case), $D$ is the room depth (128 in our case), $H$ is the room height (32 in our case), and $W$ is the room width (128). We denote a set of input tensors as $X = \{x_1, \ldots, x_N\}$ where $N$ is the number of samples.

### 3D Output Data

The goal of the surrogate model is to predict a 3D heatmap of the temperature of each voxel based on the input encoding a data center. Here, we describe the construction of these ground truth heatmaps, each of which we denote as a tensor $y \in \mathrm{R}^{1 \times D \times H \times W}$. We denote an element within the output tensor $y$ as $y_{1ijk}$ and a set of output tensors as $Y = \{y_1, \ldots, y_N\}$. Volumetric temperature heatmaps acquired through $6SigmaDCX$ simulations are sparse and irregularly sampled on all axes (see Figure 4). With 3D surrogate modeling, we are performing dense volumetric regression and, in turn, require dense ground truth targets to learn from. To make this sparse heatmap dense, the nearest interpolation is performed over a voxel grid. While linear and cubic interpolations can produce more realistic gradients, these techniques are computationally intractable for large 3D datasets when the sampled points are irregular and unstructured. To improve the realism of the ground truth heatmap interpolation, we use a Gaussian filter to smooth the data with $\sigma = 0.5$ and a radius of $4\sigma$. Last, the heatmap is truncated at the minimum and maximum temperatures of 20 °C and 70 °C, respectively. This affects less than 1% of the data and is effective at removing anomalous simulation values.

### Data Preprocessing and Training

All data $X$ inputted into the 3D surrogate model ($f$) has z-score normalization to ensure each channel possesses a mean of zero and a standard deviation of one, which standardizes the data and facilitates model training. To improve robustness and prevent overfitting, we apply data augmentation (random 90° rotations, vertical and horizontal flips) during epoch-wise shuffling. The initial learning rate was set at $1 \times 10^{-3}$, with a plateau-based learning rate scheduler employed to adjust this parameter in response to the performance on the validation set, which comprises 10% of the training data, guided by the $smoothL1$ loss metric. Training was performed for the FNO model on a single GPU with a batch size of 8, whereas all other models were trained on 8 GPUs with the same per-GPU batch size which yields an effective total batch size of 64. A weight decay of $1 \times 10^{-8}$ was applied to regulate learning and mitigate overparameterization. The parameters of $f$ are updated using the Adam optimizer. We opt to use the $smoothL1$ loss as given by:

$$L(Y, \hat{Y}) = \frac{1}{NDHW} \times \sum_{n,i,j,k=1}^{N,D,H,W} L_{smooth}(y_{nijk}, \hat{y}_{nijk})$$

where $L_{smooth}$ is defined as

$$L_{smooth}(y, \hat{y}) = \begin{cases} \frac{(y-\hat{y})^2}{2}, & \text{if } |y - \hat{y}| < \beta \\ |y - \hat{y}| - \frac{\beta}{2}, & \text{otherwise} \end{cases}$$

and $\beta$ (set to 1.0) is a hyperparameter. This loss is less susceptible to outliers than mean-squared-error (MSE) and can mitigate exploding gradients in some cases (Girshick 2015). We clip gradient updates such that the norm is less than one, which also helps prevent exploding gradients.

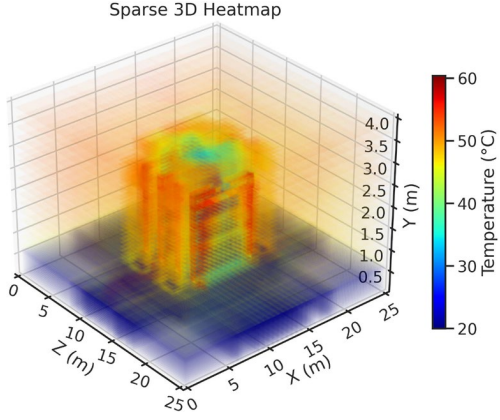

Figure 4: Heatmap for a simple data center configuration. See Figure 8 for an example of the 2D slices of a dense heatmap along the height dimension.

We considered several 3D architectures to construct the 3D surrogate model to show the effectiveness of the data modeling process for these architectures. Below, we list the architectures considered with brief descriptions:

- **3D U-Net** (Çiçek et al. 2016): An adaptation of U-Net (Ronneberger, Fischer, and Brox 2015) designed for 3D data, such as volumetric data.

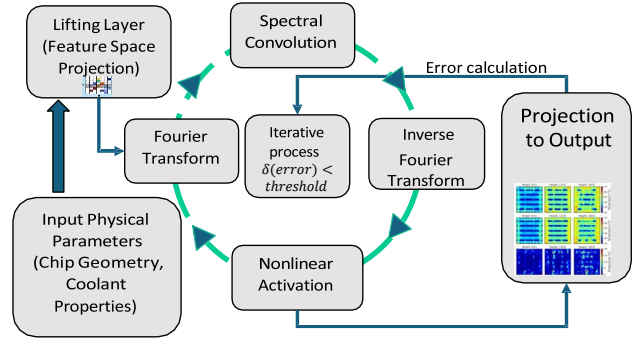

Figure 5: 3D FNO training process for datacenter

- **3D Residual-UNet** (Lee et al. 2017): Based on the concept of residual learning, this architecture introduces residual blocks into the 3D U-Net framework, enhancing its ability to learn from volumetric data by enabling the training of deeper networks.
- **3D U-Net++** (Zhou et al. 2019): A variant of the traditional U-Net architecture, UNet++ was first proposed for medical image segmentation. In this variant, the encoder and decoder sub-networks are connected through a series of nested, dense skip pathways aiming to reduce the semantic gap between the feature maps.
- **3D Residual-UNetSE** (Toubal, Duan, and Yang 2020): Enhancing the 3D Residual-UNet model, this architecture incorporates Squeeze and Excitation (SE) blocks. These blocks re-calibrate channel-wise feature responses by explicitly modeling interdependencies between channels, improving the model's sensitivity to informative features for high-resolution medical volume segmentation.
- **3D Swin UNETR** (Hatamizadeh et al. 2022): A recent variant of U-Net, "Swin UNETR", addresses the limitations of convolution-based architectures in modeling long-range information by utilizing shifted windows for computing self-attention.
- **3D Fourier Neural Operator** The FNO framework is a spectral neural architecture designed for solving parametric partial differential equations. Its core innovation lies in the application of the Fast Fourier Transform (FFT) within neural operator layers, facilitating efficient and resolution-independent learning of mappings from input conditions to PDE solution fields (Li et al. 2020; Peng et al. 2024).

Formally, given an input function $a(x)$ defined over domain $\Omega$, the FNO first lifts $a(x)$ into a high-dimensional latent space using a learned linear transformation:

$$v_0(x) = P(a(x)), \quad (1)$$

where $P$ is a trainable projection operator. Subsequent layers act globally via convolution in Fourier space:

$$v_{t+1}(x) = F^{-1}(R \cdot F(v_t(x))) + W(v_t(x)), \quad (2)$$

where $F$ and $F^{-1}$ denote the Fourier and inverse Fourier transforms, $R$ is a trainable Fourier-domain linear map,

and $W$ is a learned nonlinear local transformation. This architecture allows the FNO to efficiently capture non-local dependencies and model highly nonlinear physical dynamics.

Compared to conventional convolutional neural networks (CNNs) or graph-based neural operators, FNOs demonstrate superior resolution invariance, enabling zero-shot super-resolution—i.e., models trained at lower discretization can generalize to higher-resolution data during inference without retraining (Li et al. 2020; Peng et al. 2024). The global receptive field and efficient spectral mixing result in high accuracy for a range of PDE-driven systems, including turbulent flow and thermal propagation, at significantly reduced computational cost (Li et al. 2025).

In this study, we implement a 3D Fourier Neural Operator (FNO) as part of a suite of surrogate architectures for rapid volumetric prediction of temperature fields in data centers. The FNO is supplied with structured 3D input tensors, where each tensor encodes (i) normalized server and IT component power (proxy for heat generation), (ii) CRAC temperature setpoints, and (iii) CRAC fan speeds, each mapped to the corresponding spatial locations on the voxel grid. These inputs reflect the three critical dimensions of thermal dynamics within the facility and are tailored to preserve spatial structure and physical relevance across diverse center configurations. The output of the FNO is a dense 3D temperature heatmap, interpolated and smoothed from CFD simulations and representing the steady-state thermal profile at fine (15 cm) voxel resolution. By training on a composite dataset generated via Latin Hypercube Sampling of operational scenarios, the FNO surrogate learns to accurately approximate CFD outcomes

# Experiments and Results

## Setup

Experiments were run on 2× Intel Xeon Gold 6248 CPUs (20 cores, 40 threads) and 8× Nvidia Tesla V100 GPUs. FNO was trained on a single GPU (16 hours), while other models were trained with all 8 GPUs (2 hours).

## Evaluation Metrics

To evaluate the effectiveness of our approach, we consider several metrics: inference time, mean-squared-error (MSE), mean absolute error (AE), top-$t$ AE, and 3D structural similarity (SSIM). Top-$t$ AE is especially useful for evaluating how well our approach models hot-spots in data centers.

Each metric is defined below.

$$\text{MSE}(Y, \hat{Y}) = \sum_{n,i,j,k=1}^{N,D,H,W} \frac{(y_{nijk} - \hat{y}_{nijk})^2}{NDHW}$$

$$\text{Mean AE}(Y, \hat{Y}) = \sum_{n,i,j,k=1}^{N,D,H,W} \frac{|y_{nijk} - \hat{y}_{nijk}|}{NDHW}$$

$$\text{Top-}t\text{ AE}(Y, \hat{Y}) = \sum_{n=1}^{N} \sum_{i,j,k \in \text{top-}t(n)} \frac{|y_{nijk} - \hat{y}_{nijk}|}{Nt}$$

The function top-$t(\cdot)$ returns the 3D indices of the largest $t$ elements of its tensor argument, i.e., the hottest $t$ temperature locations of a ground truth heatmap. In our evaluation, we set $t$ to 10% of the number of heatmap elements: $t = \lceil DHW/10 \rceil$. The inference time is taken as the time to predict a single sample after the model and data are loaded into GPU memory.

## 3D Surrogate Results

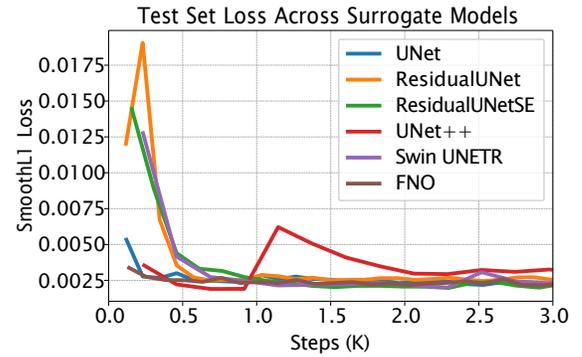

Figure 6: Evolution of the $smooth\ L1$ loss function on the test set.

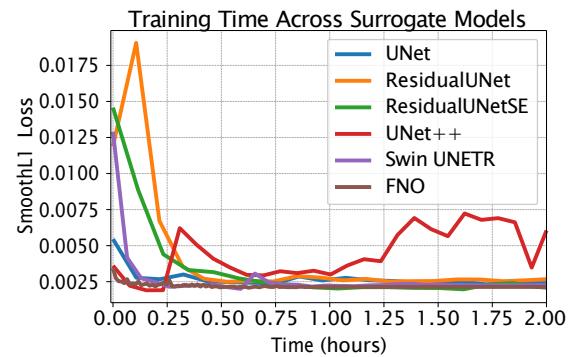

Figure 7: Training time comparison across surrogate models.

**Training Performance and Inference Speed** Table 1 and Figure 6 evaluate various surrogate models in terms of inference time, error and structural similarity metrics. The similarity in evaluation performance, evolution of the losses

| Model | Inf. Time (s) ↓ | MSE ↓ | Mean AE (°C) ↓ | Top-$t$ AE (°C) ↓ | 3D SSIM ↑ |
|---|---|---|---|---|---|
| U-Net | 0.109 | 0.00363 | 1.89 | 1.80 | 0.8171 |
| Residual-UNet | 0.127 | 0.00496 | 2.15 | 2.54 | 0.7869 |
| Residual-UNetSE | 0.114 | 0.00431 | 2.08 | 2.84 | 0.7861 |
| U-Net++ | 0.071 | 0.00525 | 5.98 | 7.29 | 0.7818 |
| Swin UNETR (Transformer) | 0.090 | 0.00483 | 2.60 | 2.10 | 0.8056 |
| Fourier Neural Operator | 0.171 | 0.00427 | 2.15 | 2.02 | 0.8260 |

Table 1: Evaluation metrics on test set. The inference time is computed on a V100 GPU with a batch size of 1.

(Figure 6) and training time (Figure 7) indicate that the data modeling approach works well across different types of surrogate model. Note that the FNO's single-GPU training time has been scaled by $1/8$ to match the 8-GPU trainings.

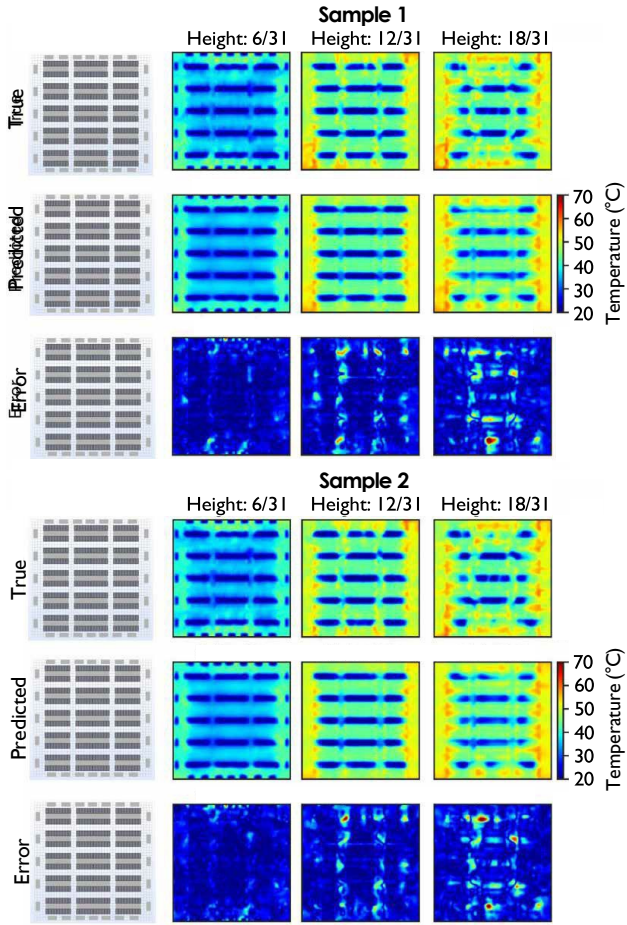

Figure 8: Matrix representation across different slices of the data center room. Two random samples. True outputs are in the top row, model predictions are in the middle row, and the error map is in bottom row. Each column represents a slice at varying heights.

In terms of inference, the models significantly enhance performance on GPUs compared to traditional CFD approaches. As seen in similar works on flow variable predictions using Finite Dimensional Operator based methods (Guo, Li, and Iorio 2016), ML surrogate models can yield substantial improvements in performance with minimal approximation error. This is evident in the recorded inference times, where even models with more complex architectures, such as Swin UNETR, provide results in a much shorter time (100 ms vs. 35 minutes), reinforcing the usefulness of ML surrogate models for simulating steady-state temperature predictions for data centers. To our knowledge, no comparable 3D thermal surrogates exist for steady-state prediction. Hence we evaluate our approach by comparing the results against the ground truth.

**Prediction Analysis** Figure 8 provides a detailed matrix comparing the true CFD generated heatmap (row 1), with the U-Net 3D model's predictions (row 2), and the resulting absolute error map ($|True - Predicted|$) (row 3) across different horizontal slices of the data center room. Each column corresponds to a horizontal slice sampled from the z axis, progressing from the bottom (height: 6/31) to a higher position (height: 18/31) of the data center room in regular intervals. We chose these intervals to showcase the instances where the deviations from true values occur, and there are more occurrences of hot spots. Other instances have more homogeneous 2D slices with almost perfect prediction.

While looking at the error heat map, we notice most of the errors occurring at the physical boundaries where the Air Cooling Units are located (height: 6/31), intermediate space between the cabinets where the cold air enters the cabinets (height: 12/31), or the cabinets are separated from each other (height: 18/31). The arrangement of these boundaries differs from those observed in the training data (see Figure 3). This implies that the models have not overfitted the particular characteristics in the training data.

### Downstream Tasks

The surrogate model accelerates tasks relevant to data center design and sustainability. Using the CFDDC model and a genetic algorithm (GA), we optimized server workload distribution to minimize temperature hotspots (Figure 9).

The CFDDC predicts thermal impacts of GA-generated arrangements. The genetic algorithm (GA) optimization led to a marked improvement in the data center's thermal management. We employed a population size of 20, a mutation rate of 1%, and a crossover probability of 90%. Illustrated in Figure 10, the "Best" line demonstrates a significant decrease in normalized maximum temperature over 25 generations, indicating the GA's success in identifying server workload distributions that minimize hotspots. The GA's ef-ficacy is highlighted by a **7.70%** reduction in hotspots compared to a non-optimized baseline that randomly distributes workload, showcasing its potential for enhancing cooling efficiency and achieving more favorable thermal conditions within the data center.

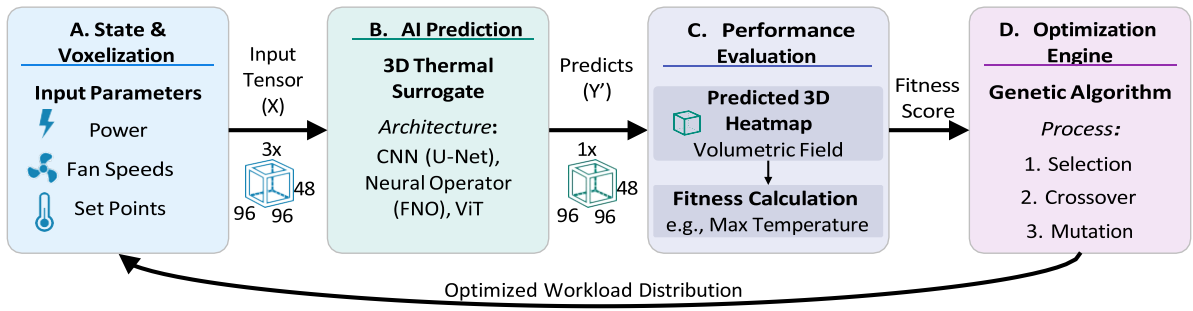

Figure 9: The closed-loop optimization framework. (A) Data center parameters are voxelized into a 3D input tensor. (B) A 3D Thermal Surrogate (e.g., U-Net, FNO, ViT) predicts a 3D heatmap. (C) A scalar fitness score (like max temperature) is extracted from the heatmap. (D) This score guides a Genetic Algorithm to propose an optimized workload distribution, completing the loop.

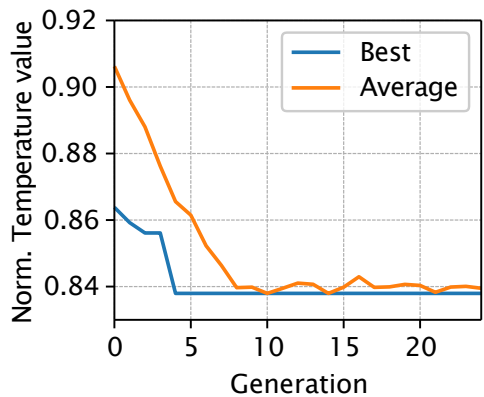

Figure 10: Evolution of the normalized temperature value over successive generations.

The overall goal of this approach in the digital twins framework is to reduce the cooling energy and the carbon footprint for processing workloads, improving data center sustainability. susdcaaa124, This fits into the earlier work on multi-agent holistic data center sustainability (Naug et al. 2024, 2023, 2025; Sarkar et al. 2024b,a, 2023b,a, 2025; Guillen-Perez et al. 2025).

### Future Work and Deployment Advantage

One big advantage of this method over traditional CFD tools is deployment efficiency and cost savings. Traditional CFD tools require costly calibration for deployment at each data center. However, unlike the CFD tool, the CFDDC surrogate can train continuously on sensor data at deployment and can converge to the characteristics of the data center. This would lead to high accuracy at a lower cost.

Even though this method relies on CFD generated data to train the 3D vision models as surrogate, this method generalizes for different sizes and configurations of DC. To address supercomputing scale we are working on a downstream stitching mechanism of the 3D heat maps from different subsections.

### Conclusions

This paper shows how to create a 3D ML thermal surrogate for CFD tools for data centers by structuring the thermal attributes such as heat sources, airflow, and temperature control as input channels. This technique is generalizable across 3D vision architectures like U-Net variants, neural operators like FNO and vision transformers like Swin UNETR. CFDDC provides accurate temperature predictions (top-10 absolute error of about 2°C) and significantly speeds up computations (20,000x) to identify hotspots, enabling real-time control and iterative design optimizations for minimizing energy usage, reducing downtime, and extending server life. This significantly impacts operating costs and sustainability by reducing the carbon footprint and e-waste. This novel thermal CFD surrogate modeling approach is generalizable and can be applied to other domains requiring steady-state analysis. We plan to extend this work to cooling control for buildings and large facilities in smart cities.

### Impact Statement

This paper advances data center sustainability by introducing a 3D thermal surrogate of a CFD model. Our approach significantly reduces computational costs and carbon footprints while enabling real-time thermal management. The societal impact includes energy-efficient data center operations and broader applicability to sustainable infrastructure and building and industrial management. We do not foresee any ethical concerns beyond standard considerations in AI deployment.

### Acknowledgments

We would like to thank Cullen Bash for his guidance on cooling technology and sustainability in data centers.

# References

6SigmaDCX — Future Facilities. 2023. CFD simulation tool. https://www.futurefacilities.com/resources/videos/products/introducing-6sigmadcx/. [Accessed 11-08-2023].

Asgari, S.; MirhoseiniNejad, S.; Moazamigoodarzi, H.; Gupta, R.; Zheng, R.; and Puri, I. K. 2021. A gray-box model for real-time transient temperature predictions in data centers. *Applied Thermal Engineering*, 185: 116319.

Athavale, J.; Yoda, M.; and Joshi, Y. 2019. Comparison of data driven modeling approaches for temperature prediction in data centers. *International Journal of Heat and Mass Transfer*, 135: 1039–1052.

Girshick, R. 2015. Fast R-CNN. In *Proceedings of the IEEE international conference on computer vision*, volume abs/1504.08083, 1440–1448.

Guibas, J.; Mardani, M.; Li, Z.; Tao, A.; Anandkumar, A.; and Catanzaro, B. 2021. Adaptive Fourier Neural Operators: Efficient Token Mixers for Transformers.

Guillen-Perez, A.; Naug, A.; Gundecha, V.; Ghorbanpour, S.; Gutierrez, R. L.; Babu, A. R.; Salim, M.; Banerjee, S.; Essink, E. H. O.; Fay, D.; et al. 2025. DCcluster-Opt: Benchmarking Dynamic Multi-Objective Optimization for Geo-Distributed Data Center Workloads. *arXiv preprint arXiv:2511.00117*.

Guo, X.; Li, W.; and Iorio, F. 2016. Convolutional neural networks for steady flow approximation. In *Proceedings of the 22nd ACM SIGKDD international conference on knowledge discovery and data mining*, 481–490.

Hatamizadeh, A.; Nath, V.; Tang, Y.; Yang, D.; Roth, H.; and Xu, D. 2022. Swin UNETR: Swin Transformers for Semantic Segmentation of Brain Tumors in MRI Images. *arXiv preprint arXiv:2201.01266*.

Ilager, S.; Ramamohanarao, K.; and Buyya, R. 2020. Thermal prediction for efficient energy management of clouds using machine learning. *IEEE Transactions on Parallel and Distributed Systems*, 32(5): 1044–1056.

Jin, S.-Q.; Li, N.; Bai, F.; Chen, Y.-J.; Feng, X.-Y.; Li, H.-W.; Gong, X.-M.; and Tao, W.-Q. 2023. Data-driven model reduction for fast temperature prediction in a multi-variable data center. *International Communications in Heat and Mass Transfer*, 142: 106645.

Lee, K.; Zung, J.; Li, P. H.; Jain, V.; and Seung, H. S. 2017. Superhuman Accuracy on the SNEMI3D Connectomics Challenge. *ArXiv*, abs/1706.00120.

Li, J.; Tian, J.; Lin, Y.; Zhou, Z.; Li, Y.; Gao, B.; Tang, J.; Chen, J.; He, Y.; Qian, H.; et al. 2025. Memristive floating-point Fourier neural operator network for efficient scientific modeling. *Science Advances*, 11(25): eadv4446.

Li, Z.; Kovachki, N.; Azizzadenesheli, K.; Liu, B.; Bhattacharya, K.; Stuart, A.; and Anandkumar, A. 2020. Fourier Neural Operator for Parametric Partial Differential Equations. *arXiv*.

Lin, J.; Lin, W.; Lin, W.; Wang, J.; and Jiang, H. 2022. Thermal prediction for Air-cooled data center using data Driven-based model. *Applied Thermal Engineering*, 217: 119207.

McKay, M. D.; Beckman, R. J.; and Conover, W. J. 1979. A comparison of three methods for selecting values of input variables in the analysis of output from a computer code. *Technometrics*, 21(2): 239–245.

Morozova, N.; Trias, F. X.; Capdevila, R.; Schillaci, E.; and Oliva, A. 2022. A CFD-based surrogate model for predicting flow parameters in a ventilated room using sensor readings. *Energy and Buildings*, 266: 112146.

Naug, A.; Guillen, A.; Kumar, V.; Greenwood, S.; Brewer, W.; Ghorbanpour, S.; Babu, A. R.; Gundecha, V.; Gutierrez, R. L.; and Sarkar, S. 2025. LC-Opt: Benchmarking Reinforcement Learning and Agentic AI for End-to-End Liquid Cooling Optimization in Data Centers. *arXiv preprint arXiv:2511.00116*.

Naug, A.; Guillen, A.; Luna, R.; Gundecha, V.; Bash, C.; Ghorbanpour, S.; Mousavi, S.; Babu, A. R.; Markovikj, D.; Kashyap, L. D.; Rengarajan, D.; and Sarkar, S. 2024. SustainDC: Benchmarking for Sustainable Data Center Control. In *Advances in Neural Information Processing Systems*, volume 37, 100630–100669. Curran Associates, Inc.

Naug, A.; Guillen, A.; Luna Gutiérrez, R.; Gundecha, V.; Ghorbanpour, S.; Dheeraj Kashyap, L.; Markovikj, D.; Krause, L.; Mousavi, S.; Babu, A. R.; and Sarkar, S. 2023. PyDCM: Custom Data Center Models with Reinforcement Learning for Sustainability. In *Proceedings of the 10th ACM International Conference on Systems for Energy-Efficient Buildings, Cities, and Transportation*, BuildSys '23, 232–235. New York, NY, USA: Association for Computing Machinery. ISBN 9798400702303.

Peng, W.; Qin, S.; Yang, S.; Wang, J.; Liu, X.; and Wang, L. L. 2024. Fourier neural operator for real-time simulation of 3D dynamic urban microclimate. *Building and Environment*, 248: 111063.

Ronneberger, O.; Fischer, P.; and Brox, T. 2015. U-net: Convolutional networks for biomedical image segmentation. In *Medical Image Computing and Computer-Assisted Intervention–MICCAI 2015: 18th International Conference, Munich, Germany, October 5-9, 2015, Proceedings, Part III 18*, 234–241. Springer.

Sarkar, S.; Naug, A.; Guillen, A.; Gundecha, V.; Luna Gutiérrez, R.; Ghorbanpour, S.; Mousavi, S.; Ramesh Babu, A.; Rengarajan, D.; and Bash, C. 2025. Hierarchical Multi-Agent Framework for Carbon-Efficient Liquid-Cooled Data Center Clusters. *Proceedings of the AAAI Conference on Artificial Intelligence*, 39(28): 29694–29696.

Sarkar, S.; Naug, A.; Guillen, A.; Gutierrez, R. L.; Gundecha, V.; Ghorbanpour, S.; Mousavi, S.; and Babu, A. R. 2023a. Sustainable Data Center Modeling: A Multi-Agent Reinforcement Learning Benchmark.

Sarkar, S.; Naug, A.; Guillen, A.; Luna, R.; Gundecha, V.; Ramesh Babu, A.; and Mousavi, S. 2024a. Sustainability of Data Center Digital Twins with Reinforcement Learning. *Proceedings of the AAAI Conference on Artificial Intelligence*, 38(21): 23832–23834.

Sarkar, S.; Naug, A.; Luna, R.; Guillen, A.; Gundecha, V.; Ghorbanpour, S.; Mousavi, S.; Markovikj, D.; and

Ramesh Babu, A. 2024b. Carbon Footprint Reduction for Sustainable Data Centers in Real-Time. *Proceedings of the AAAI Conference on Artificial Intelligence*, 38(20): 22322–22330.

Sarkar, S.; Naug, A.; Luna Gutierrez, R.; Guillen, A.; Gundecha, V.; Ramesh Babu, A.; and Bash, C. 2023b. Real-time Carbon Footprint Minimization in Sustainable Data Centers with Reinforcement Learning. In *NeurIPS 2023 Workshop on Tackling Climate Change with Machine Learning*.

Toubal, I. E.; Duan, Y.; and Yang, D. 2020. Deep Learning Semantic Segmentation for High-Resolution Medical Volumes. In *2020 IEEE Applied Imagery Pattern Recognition Workshop (AIPR)*, 1–9.

Xiao, X.; Cao, D.; Yang, R.; Gupta, G.; Liu, G.; Yin, C.; Balan, R.; and Bogdan, P. 2023. Coupled Multiwavelet Neural Operator Learning for Coupled Partial Differential Equations. *arXiv*.

Zhou, Z.; Siddiquee, M. M. R.; Tajbakhsh, N.; and Liang, J. 2019. UNet++: Redesigning Skip Connections to Exploit Multiscale Features in Image Segmentation. *IEEE Transactions on Medical Imaging*.

Çiçek, Ö.; Abdulkadir, A.; Lienkamp, S. S.; Brox, T.; and Ronneberger, O. 2016. 3D U-Net: Learning Dense Volumetric Segmentation from Sparse Annotation. In *International Conference on Medical Image Computing and Computer-Assisted Intervention*, 424–432. Springer.